\documentclass[runningheads]{llncs}
\usepackage{graphicx,xcolor}
\usepackage{comment}
\usepackage{caption}
\usepackage{amsmath, amssymb}
\begin{document}
\title{Effect of Text Color on Word Embeddings}
\author{
Masaya Ikoma\and Brian Kenji Iwana\orcidID{0000-0002-5146-6818}\and\\
Seiichi Uchida\orcidID{0000-0001-8592-7566}
}
\authorrunning{Ikoma et al.}
\institute{Kyushu University, Fukuoka, Japan\\
\email{\{masaya.ikoma, brian\}@human.ait.kyushu-u.ac.jp,\\
uchida@ait.kyushu-u.ac.jp}
}
\maketitle 

\begin{abstract}
In natural scenes and documents, we can find the correlation between a text and its color. For instance, the word, ``hot,'' is often printed in red, while ``cold'' is often in blue. This correlation can be thought of as a feature that represents the semantic difference between the words. Based on this observation, we propose the idea of using text color for word embeddings. While text-only word embeddings (e.g. word2vec) have been extremely successful, they often represent antonyms as similar since they are often interchangeable in sentences. In this paper, we try two tasks to verify the usefulness of text color in understanding the meanings of words, especially in identifying synonyms and antonyms. First, we quantify the color distribution of words from the book cover images and analyze the correlation between the color and meaning of the word. Second, we try to retrain word embeddings with the color distribution of words as a constraint. By observing the changes in the word embeddings of synonyms and antonyms before and after re-training, we aim to understand the kind of words that have positive or negative effects in their word embeddings when incorporating text color information.
\end{abstract}

\section{Introduction\label{sec:introduction}}
\begin{figure}[t]
	\begin{center}
	    \includegraphics[width=0.8\textwidth]{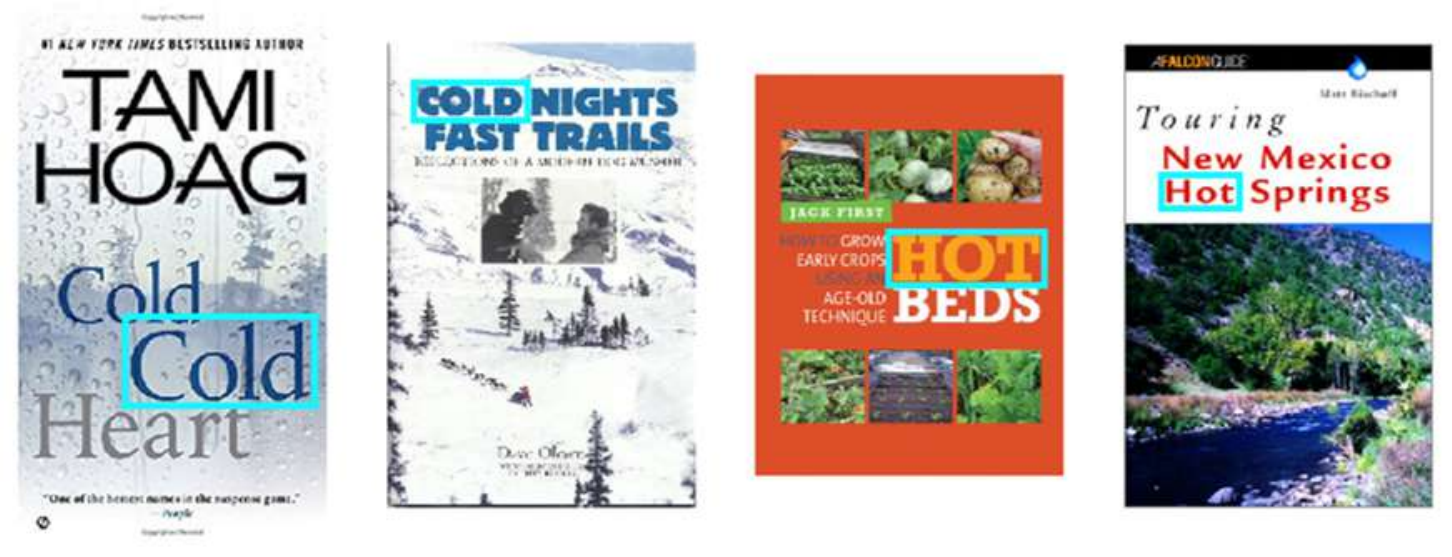}
		\caption{Example of antonyms printed in different colors.}
		\label{fig:antonyms-example}
	\end{center}
\end{figure}

In natural scenes and documents, the color of text can have a correlation to the meaning of the text. 
For example, as shown in Fig.~\ref{fig:antonyms-example}, the word, ``hot,'' is often printed in red and the word, ``cold,'' is often blue. 
This correlation is caused by various reasons. 
The color of the object described by a text will be a reason; for example, words relating to plants are often printed in green. 
Visual saliency (i.e., visual prominence) is another
reason; for example, cautions and warnings are often printed in red or yellow for a higher visual saliency. 
Another important reason is the {\em impression} of color; for example, red gives an excitement impression and pink gives a feminine impression, according to color psychology~\cite{Elliot_2014}.  \par

\begin{figure}[t]
	\begin{center}
		\includegraphics[width=0.8\textwidth]{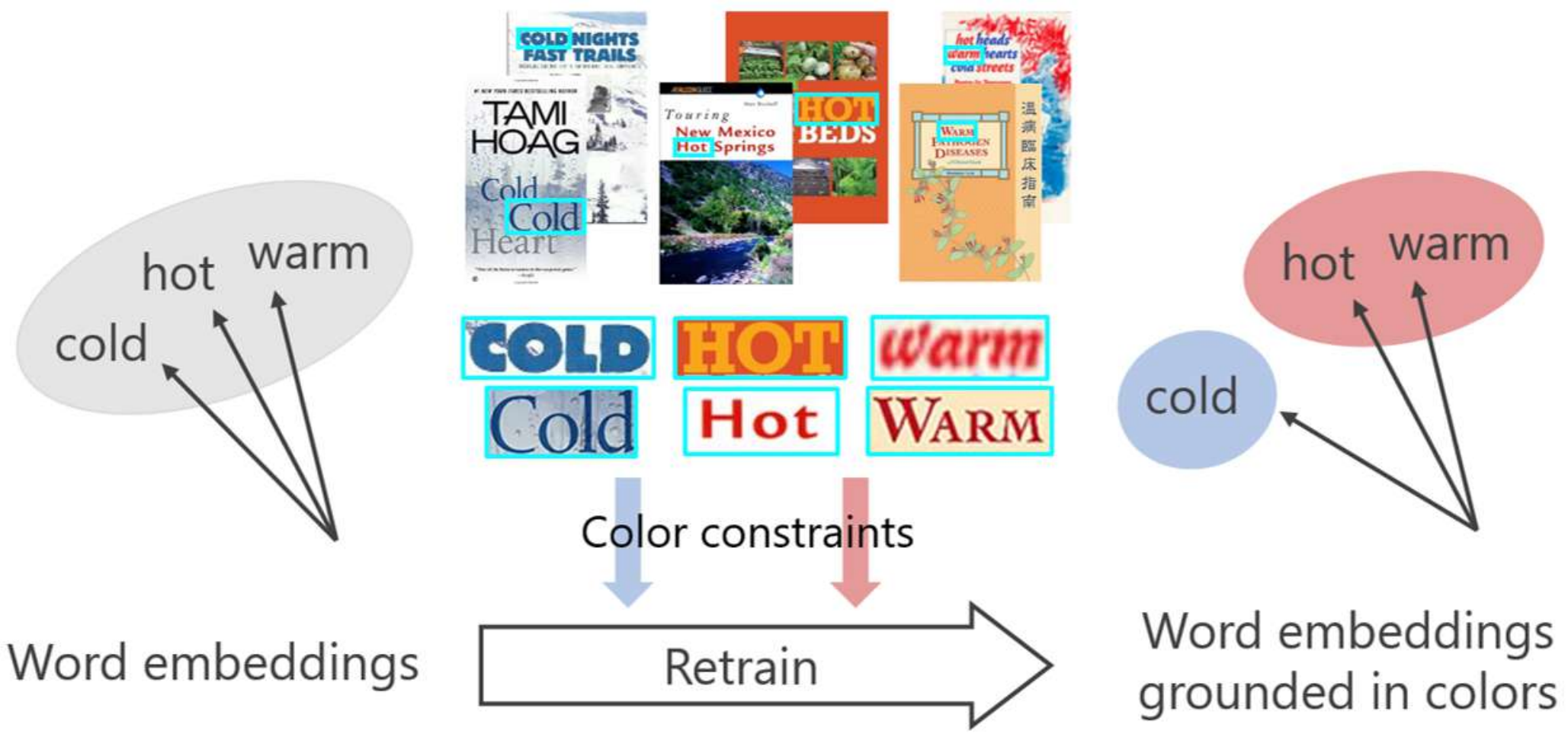}
		\caption{The concept of this paper. Correlation between colors
	 and meanings of individual words {\em might} be helpful to give better
	 semantic vectors. }
		\label{fig:concept}
	\end{center}
\end{figure}

Based on the above observations, one might imagine that the correlation between a word and its color is useful for {\em word embeddings}. 
Fig.~\ref{fig:concept} demonstrates this concept. 
Word embedding is the technique to convert a word into a
vector that represents the meaning of the word.  
Word2vec~\cite{word2vec1,word2vec2} is a standard and popular technique to create word embeddings. 
In addition, recently, BERT~\cite{devlin2018bert} and its variants have become popular. 
Although those word embedding methods have made large progress in natural language processing (NLP), further improvement is still necessary. 
For example, word2vec often gives similar vectors for antonyms (such as hot and cold) since they are often grammatically interchangeable in sentences. 
We, therefore, expect that color usage would be helpful to have better vector representations for such cases, as shown in Fig.~\ref{fig:concept}.  
\par

However, we must note that the correlation between a word and its color is not always strong. 
For example, the word, ``food,'' will be printed in various colors depending on the type of food. 
Moreover, the color of the word, ``the,'' will not have any specific trend, except for the general trend that words are often printed in achromatic color~\cite{gao2015true}. 
\par

The purpose of this paper is to understand what kinds of words have a positive or negative or no effect in their vector representation by the incorporation of color information. 
For this purpose, we first determine the color distribution (i.e., color histogram) of each word by using word images collected from book titles of about 210,000 book cover images. 
Since book cover images (especially titles) are carefully created by professional designers, we can expect that the word images from them are more helpful to find the meaningful correlation between a word and its color than word images with thoughtless color usage. 
We then analyze the color histograms to find the words whose color histogram is very different from the standard one. 
\par 

After those observations, we develop a new word-embedding method that can reflect the color usage of each word. 
On one hand, we expect positive effects using the colors; for example, antonyms with contrasting color usages, like ``hot'' and ``cold,''  will have more discriminative semantic vectors (i.e., vectors that represent the meanings of the words) than the standard word embeddings. 
On the other hand, we expect negative effects, since there are cases with no strong word-color correlation. 
Accordingly, we observe the improved or degraded cases from the standard word2vec embeddings.\par 

The main contributions of this paper are summarized as follows:
\begin{itemize}
    \item To the authors' best knowledge, this is the first trial of an objective analysis of the correlation between color and meaning of words, by using a large amount of real word images.
    \item We discover words and word classes whose color usage is very particular.
    \item We proposed a novel word-embedding method that reflects the color usage of the word. If a pair of words have similar
    color usages, their semantic vectors become more similar through the proposed method.
    \item Since synonyms and antonyms do not always have similar and dissimilar color usages, the effect of word-embedding was not always positive. However, we also revealed that many synonyms and antonyms have positive effects from the proposed word-embedding method.
\end{itemize}

\section{Related Work}\label{sec:related-work}
Word2vec is one of the most famous modern word embedding methods~\cite{word2vec1,word2vec2}. 
For word2vec, two types of models, Continuous-Bag-of-Words (CBoW) and Skip-gram, have been proposed.
In both models, it is assumed that words with similar meanings in sentences are used in similar contexts. In other words, interchangeable words in a sentence can be said to have similar meanings. 
Word2vec uses this assumption to derive a semantic vector of the word by using a neural network. \par

Although word2vec gives reasonable semantic vectors for most words, it still have a drawback caused by the above assumption. Considering two sentences, ``I drink cold water'' and ``I drink hot water'',
the antonym words, ``cold'' and ``hot'' are interchangeable and will have similar semantic vectors
in spite of their opposite meanings. Similar situations often happen for (especially, adjective and adverb) antonyms and their semantic difference is underestimated.
\par
%
To make semantic vectors more discriminative, a lot multimodal word embedding methods have been proposed to capture word semantics from multiple modalities.
%
For example, Multimodal Skip-gram (MMSG)~\cite{lazaridou2015combining}. 
In normal Skip-gram, the co-occurrence of words in sentences is used to predict the context of a word. 
However, in MMSG, in addition to the co-occurrence of words, image features related to the word are simultaneously used. 
This makes it possible to generate word embeddings by taking into account the visual information of the object. 
Similarly, Sun et al.~\cite{sun2019vcwe} propose Visual Character-Enhanced Word Embeddings (VCWE) that extract features related to the composition of Chinese character shapes and applies them to a Skip-gram framework. 
A Chinese word is composed of a combination of characters containing a wealth of semantic information, and by adding those shape-features to the word embedding, a richer Chinese word expression can be obtained.
\par
%
In addition to directly using features from modalities outside the text, a method using some ``surrogate labels'' has also been proposed.
Visual word2vec~\cite{vw2v} uses embedded expressions based on the visual similarity between words. 
In Visual word2vec, first, cartoon images with captions are clustered based on the accompanying image features as meta information. 
Next, using the obtained cluster labels as surrogate labels, the word-embeddings are retrained. 
During this, the label of the cluster to which the image belongs is predicted from the words in the caption of the image using a CBoW framework. 
As a result, words belonging to the same cluster have their similarity increased. 
For example, words with high visual relevance, such as ``eat'' and ``stare,'' have their embedded expressions changed to more similar values. 
A similar technique can be used to learn word-embeddings based on any modality. 
In Sound-word2vec~\cite{sw2v}, surrogate labels obtained by clustering the sounds are predicted from word tags, thereby acquiring word embedding that takes into account relevance in sounds. 
\par
%
In addition, there are methods to acquire multimodal word embeddings based on Autoencoders. In Visually Enhanced Word Embeddings~(ViEW)~\cite{view}, an autoencoder is used to encode linguistic representations into an image-like visual representation encoding. 
Next, a hidden layer vector is used as a multimodal representation of the word. 
\par
The purpose of this paper is to investigate the relationship between words observed in real environments and their colors and to analyze how the color information can affect the recognition of word semantics. Although there were several data-driven trials on font color usages~\cite{gao2015true,ShinaharaICDAR2019}, there is no trial to correlate the font color usage to the meaning of words, to the authors' best knowledge. Throughout this trial, we first observe the color usages on words and word categories, then propose a word-embedding method with color information, and finally reveal how the font color usage is useful and not useful for word embedding.

\section{Collecting Word Images from Book Covers\label{sec:dataset}}
The book covers sourced for the experiment are from the Book Cover Dataset\footnote{\tt
https://github.com/uchidalab/book-dataset}~\cite{iwana2017judging}. 
This dataset consists of 207,572 book cover images from Amazon Inc. 
Fig.~\ref{fig:antonyms-example} shows some example book cover images. 

In order to extract the words from the book cover images, the Efficient and Accurate Scene Text detector~(EAST)~\cite{east} is used. 
EAST is a popular text detector that implements a Fully-Convolutional Network~(FCN)~\cite{long2015fully} combined with locality-aware Non-Maximum-Suppression~(NMS) to create multi-oriented text boxes. 
After detecting the words using EAST, a Convolutional Recurrent Neural Network~(CRNN)~\cite{crnn} is used to recognize the text boxes into text strings. 
Note that book title information is given as the meta-data of the Book Cover Dataset and thus it is possible to remove the words that are not contained in the book title. 
Consequently, we collected 358,712 images of words. 
Their vocabulary size (i.e., the number of different words) was 18,758. 
Each word has 19 images on average, 4,492 at maximum, and 1 at minimum. 
Note that stop words, proper nouns, digits, compounded words (e.g., ``don't'' and ``all-goals'') are removed from the
collection\footnote{From 579,240 images of words given by EAST, 492,363 images remain after removing non-title words and misrecognized words. Finally, 358,712 images remain after removing stop words, etc.}. 
In addition, we identify word variants (e.g., ``meets'' = ``meet''), by using a lemmatizer\footnote{\tt www.nltk.org/\_modules/nltk/stem/wordnet.html}.

\section{Color Usage of a Word as a Histogram}
In this paper, we represent the color of a word in the CIELAB color space. 
Each color is represented by a three-dimensional vector $(L^*, a^*, b^*)$, where $L^*$ is lightness ($\sim$intensity) and $a^*$ and $b^*$ are color channels. 
The advantage of using CIELAB is that the Euclidean distance in the CIELAB color space is approximately relative to human's perceptual difference.\par 

CIELAB color values of each word in the detected bounding box are determined as follows. 
The first step is the separation of the character pixels from the background pixels. 
Fortunately, most word images on book covers have a high contrast between the characters and the background (possibly due to better legibility). 
Therefore, a simple Otsu binarization~\cite{Otsu_1979} technique was enough for the separation.
As the second step, the CIELAB color vector $(L^*, a^*, b^*)$ is determined for a word by taking the average of all character pixels.\par 


In order to analyze the distribution of colors, we implement a color histogram.
Specifically, all colors are quantized into one of 13 basic colors (pink, red, orange, brown, yellow, olive, yellow-green, green, blue, purple, white, gray, and black), which are defined by as basic colors in the ISCC-NBS system.  
The color quantization is performed using the nearest neighbor rule with Euclidean distance in CIELAB space. 
If a word occurs $P$ times, in the dataset, we initially have a color histogram with $P$ votes for $K=13$ bins. 
We then normalize it to be its total votes equal to one, in order to remove the difference of $P$.
\par

In the following analysis, we introduce the following two conditions. 
First, we do not use the word images whose word color is achromatic (i.e., white or black or gray) {\em and} background color is chromatic (i.e., one of the 10 chromatic colors). 
A survey of color usages of scene texts and their background~\cite{gao2015true} proved that if the background color is chromatic and the foreground text is achromatic then the designer typically uses the background color to portray meaning. 
We, therefore, extend this idea to book cover title colors and remove the word images in this case. 
Note that if both title and background colors are achromatic, we do not remove the image. 
For the second condition, we removed the histogram for words which occur less than five times to limit the analysis to reliable color usages.\par

\begin{figure}[t]
	\begin{center}
		\includegraphics[width=\textwidth]{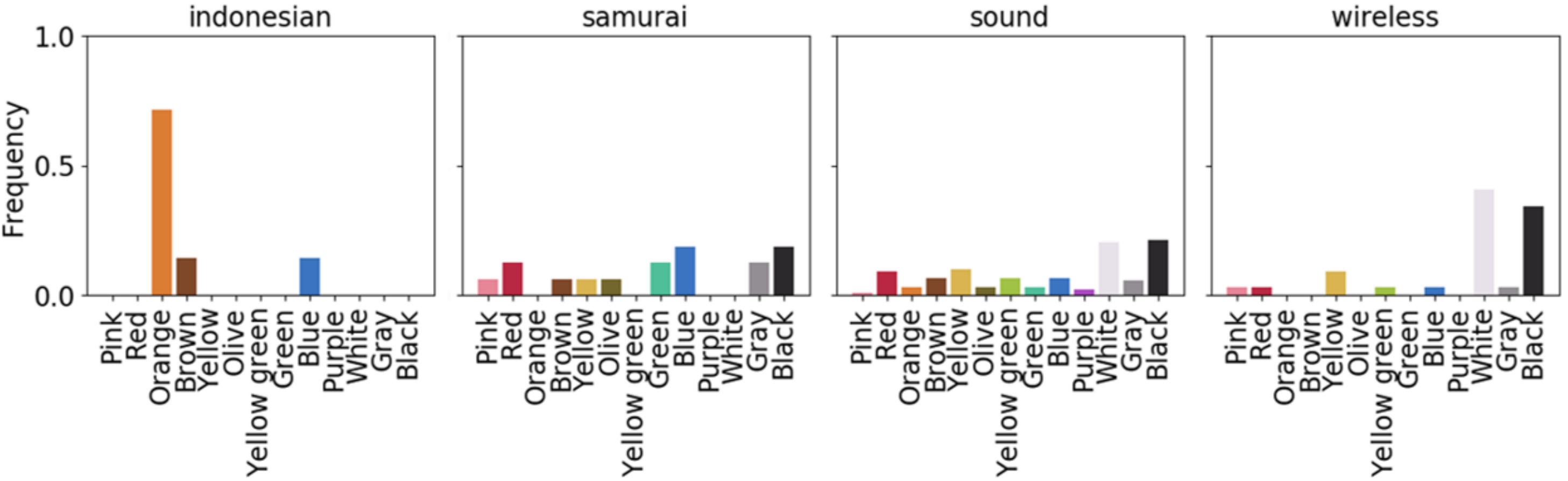}\\[-2mm]
		\caption{Color histogram examples.\label{fig:example-color-histogram}}
		\medskip
 	\end{center}
 \end{figure}

Consequently, we have color histograms of 6,762 words.  Fig.~\ref{fig:example-color-histogram} shows several examples, where each bar of the color histogram is painted by the corresponding basic color. 
The leftmost histogram, ``indonesian,'' shows a peak on a specific color, ``Orange,''  whereas the second and the third show rather flat usages. 
As shown in the rightmost histogram, we sometimes find histograms with frequent use of achromatic colors.
\par

%

\section{Words with Particular Color Usages}\label{sec:color-dist}

\subsection{How to Find Words with Particular Color Usage}

\begin{figure}[t]
	\begin{center}
		\includegraphics[width=0.65\textwidth]{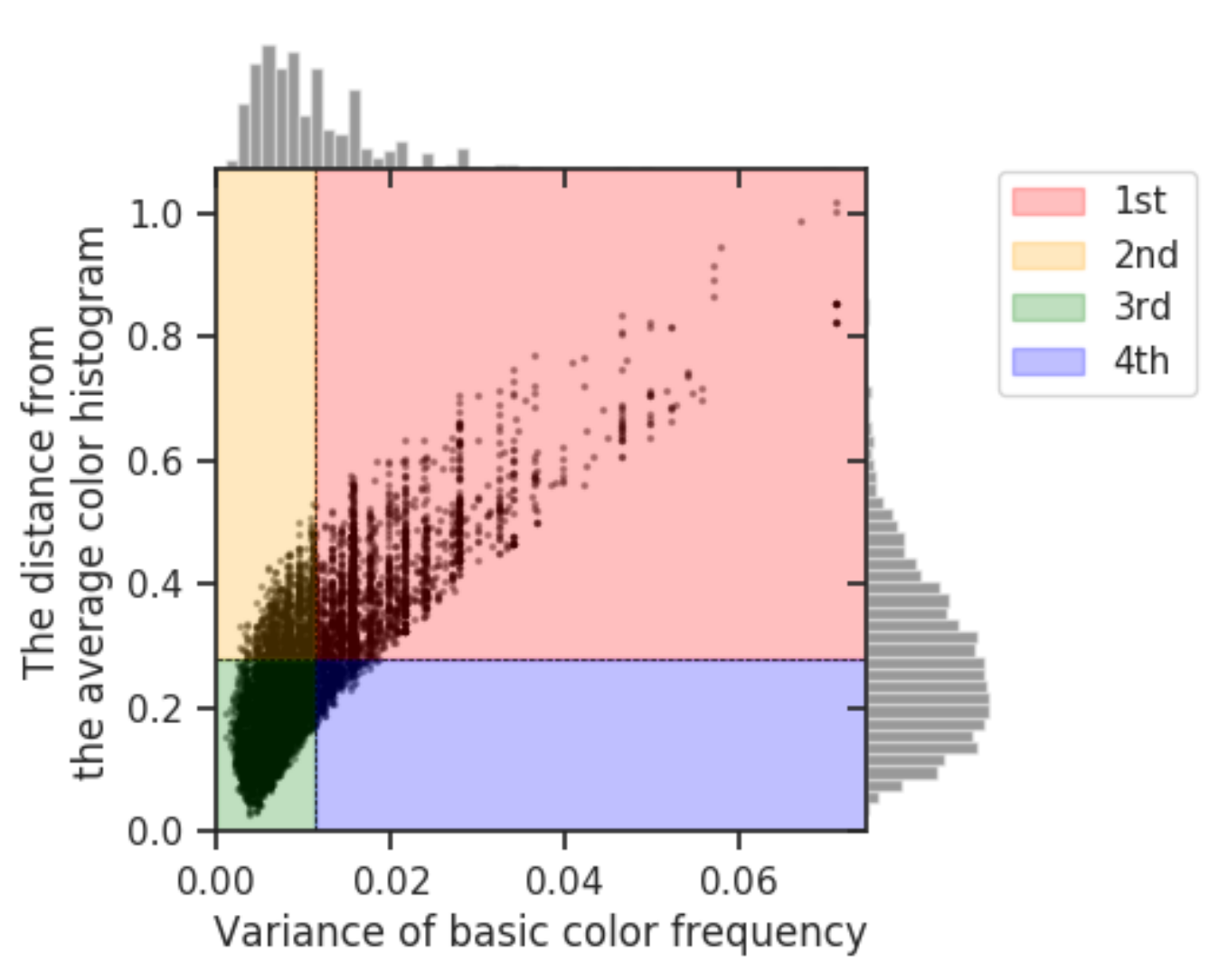}\\[-2mm]
		\caption{The distribution of 6,762 words in a two-dimensional plane evaluating the particularity of the color usage of the word.\label{fig:scat-hist}}
	\end{center}
\end{figure}

Our interest is to find the words with particular color usage rather than words with standard color usage because particular color usage might show specific correlations between the meanings and the color of the word. 
Fig.~\ref{fig:scat-hist} shows a scatter plot for finding such words. 
Each dot corresponds to a word and the horizontal axis is the color variance $\sum_k (h_w^k - \mu)^2/K$, where $h_w=(h_w^1,\ldots,h_w^K)$ is the (normalized) color histogram of the word $w\in V$ and $\mu=\sum_kh_w^k/K=1/K$. 
When this variance is large for a word $w$, the histogram is not flat but has some peaks at specific colors; that is, the word $w$ has a more particular color usage.
The vertical axis is the distance from the average color histogram $\bar{h}$, i.e., $\|h_w - \bar{h}\|$, where $\bar{h}=\sum_w h_w/|V|$, where $|V|$ is the vocabulary size. 
When this distance is large for word $w$, the histogram $h_w$ is largely deviated from the average color histogram $\bar{h}$ and word $w$ has a more particular color usage. 
Note again, $K=13$. 
\par

By splitting each axis by its average, we have four regions in Fig.~\ref{fig:scat-hist}. 
Among them, we call the region with red color {\em the first quadrant}. 
Similarly, orange, green, and blue regions as the second, third, and fourth quadrants, respectively. 
Each word belonging to the first quadrant will have a particular color usage. 
In contrast, the words in the third quadrant will have a flat and standard color usage. 
The first, second, third, and fourth quadrants contain 1,973, 1,081, 3,381, and 327 words, respectively. 
This suggests that words with particular color usages are not dominant but exist
to a certain amount.  
Specifically, although $3,381/6,762\sim 50\%$ words have an average and flat color usage and their color usage is not reflecting their meanings, the remaining $50\%$ words have a particular color usage which might reflect their meanings. 
In addition, the first quadrant with $1,973/6,762\sim 30\%$ words have a very particular color usage.

\begin{figure}[t]
	\begin{center}
		\includegraphics[width=\textwidth]{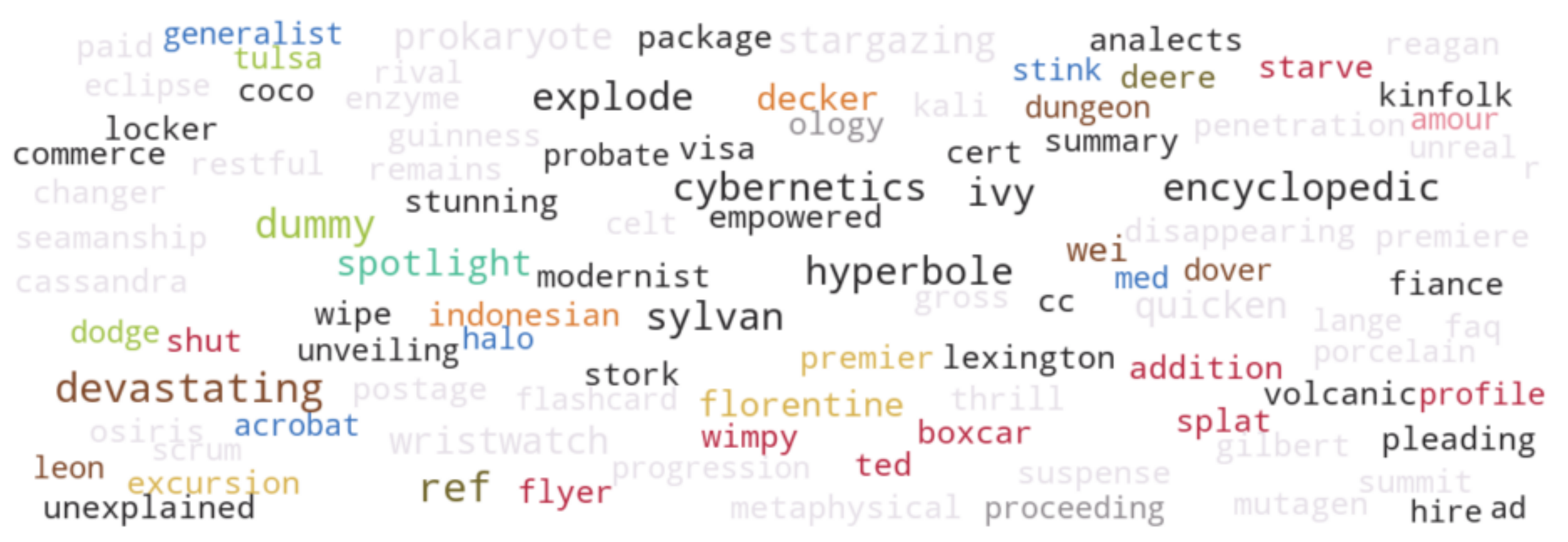}
		\caption{
		Examples of 100 words in the  first quadrant that have a particular color usage.
		\label{fig:wordcloud} 
		}
		\end{center}
\end{figure}

\subsection{Words with Particular Color Usage}
Fig.~\ref{fig:wordcloud} shows a word-cloud of the words in the first quadrant of Fig.~\ref{fig:scat-hist}, colored by each word's most frequent basic color. 
Different from typical word-clouds, the size of the word is proportional to the particularity of the word, which is calculated as the product of ``the distance from the average color histogram'' and ``variance of basic color frequency'' in Fig.~\ref{fig:scat-hist}.  
As a general tendency, achromatic colors are the most frequent color for most words and it can be seen that the meaning and color of words are not necessarily related. 
On the other hand, some relationships between the color and the meaning of words can be found. 
For example, ``premier'' is reminiscent of position and victory and is yellow while ``amour'' means love and frequently appears in pink, etc. 

\subsection{Word Category with Particular Color Usages}

\begin{figure}[t]
	\begin{center}
		\includegraphics[width=\textwidth]{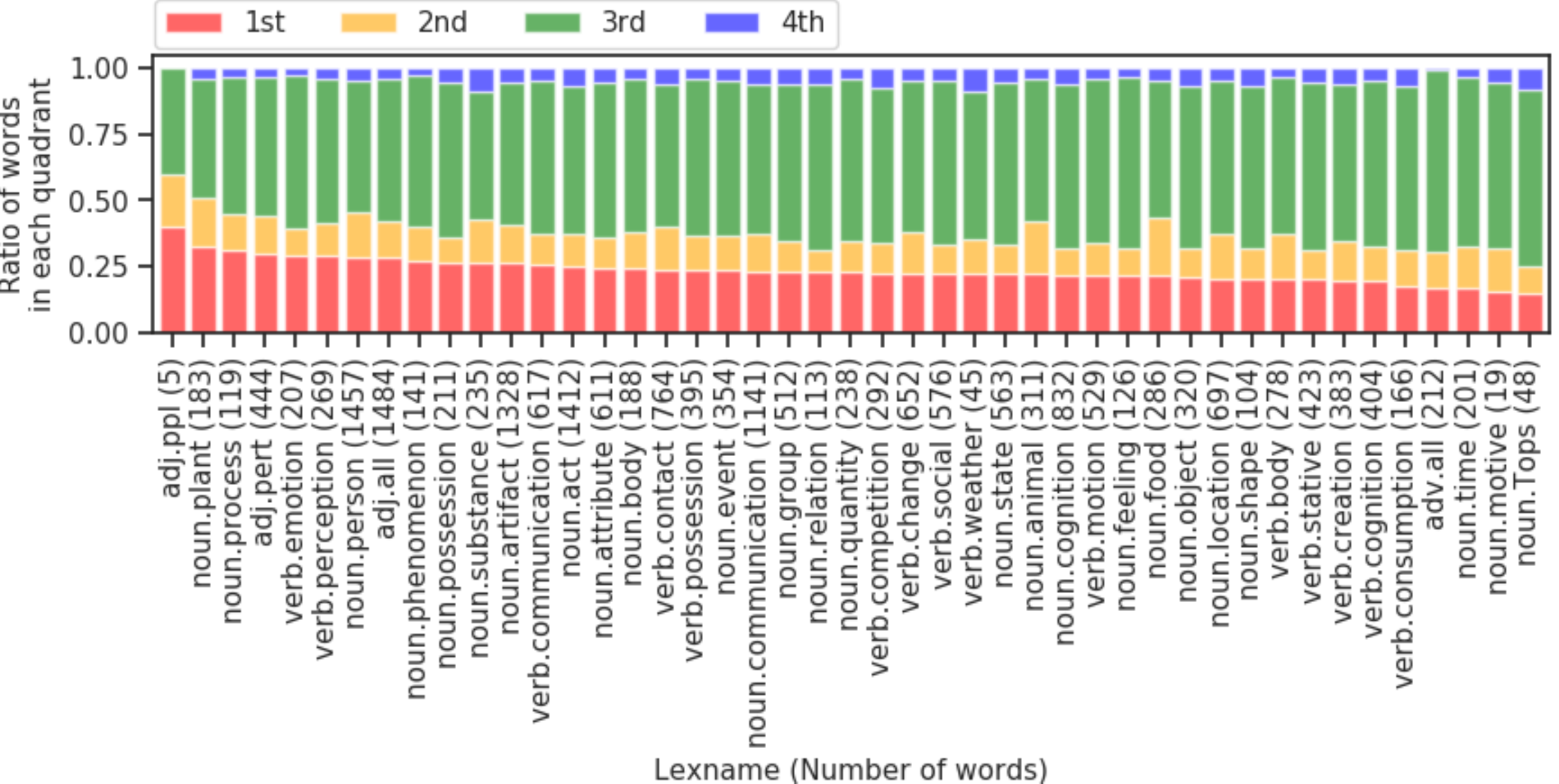}
		\caption{
		The ratio of words in each quadrant by word category (i.e., lexnames).}
		\label{fig:analy-category}
	\end{center}
\end{figure}

Words can be categorized into hierarchical groups, thus, in this section, we analyze the relationship between word category and color usage. 
Fig.~\ref{fig:analy-category} shows the ratio of the four quadrants for 45 word categories, which are defined as {\em lexnames}\footnote{\tt wordnet.princeton.edu/documentation/lexnames5wn} in WordNet. 
In this figure, the categories are ranked by the ratio words in the first quadrant.
Note that a word can belong to multiple categories by the hierarchical structure of WordNet.\par

\begin{figure}[t]
    \begin{center}
    \includegraphics[width=\textwidth]{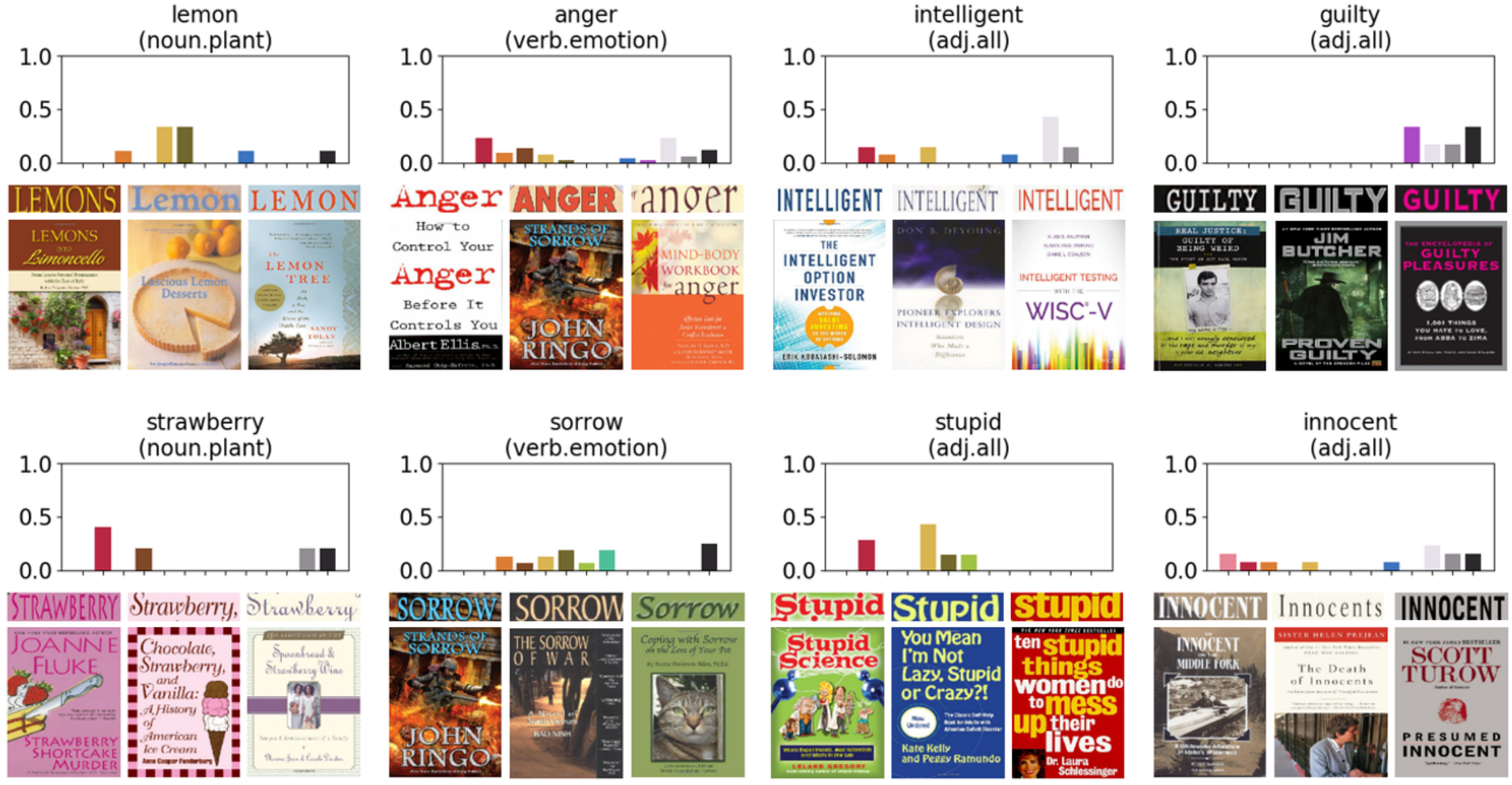}
    \caption{ 
    Words with their color histogram and image examples.
     The parenthesized term (e.g., ``noun.plant'') is the category of the word.
     }
    \label{fig:catword}
    \end{center}
\end{figure}

The category ``noun.plant'' is high-ranked; namely, words relating to plants have particular color usages. 
This is because words related to plants tend to appear in green or their fruit color. 
Fig.~\ref{fig:catword} shows the color histograms of ``lemon'' and ``strawberry'' and sample books of the dominant colors. 
In particular, ``lemon'' often appears as yellow or olive and ``strawberry'' frequently is in red. 
It is noteworthy that the rank of ``noun.plant'' is much higher than the rank of ``noun.food.'' 
This is because of the large variations of foods which result in the large variations of color usages.\par

The category ``verb.emotion'' is also high-ranked. 
As shown in Fig.~\ref{fig:catword}, the color histograms of ``anger'' shows a high frequency of warm reddish colors. 
Compared to ``anger'', the word ``sorrow'' is not printed with red but more with pale colors. 
This may be because the impression of colors affects the word color usages.  \par

It should be emphasized that categories of adjectives such as ``adj.all'' (all
adjective clusters) and ``adj.pert'' (relational adjectives, i.e., so-called pertainyms) are also high-ranked. 
These categories contain many antonym adjective pairs, like ``hot'' and ``cold,'' as well as synonym adjective pairs. 
Fig.~\ref{fig:catword} also shows the color histograms of pairs of antonyms, ``intelligent'' and ``stupid'', and ``guilty'' and ``innocent''. 
The word, ``intelligent,'' tends to be printed in paler color than ``stupid.''  
The word, ``guilty,'' was printed in purple more than ``innocent'' and did not show up as a warm color.
\par


\section{Word Embeddings Grounded in Color\label{sec:embedding}}

\subsection{Methodology\label{sec:methodology}}

\begin{figure}[t]
\begin{center}
	\includegraphics[width=0.8\textwidth]{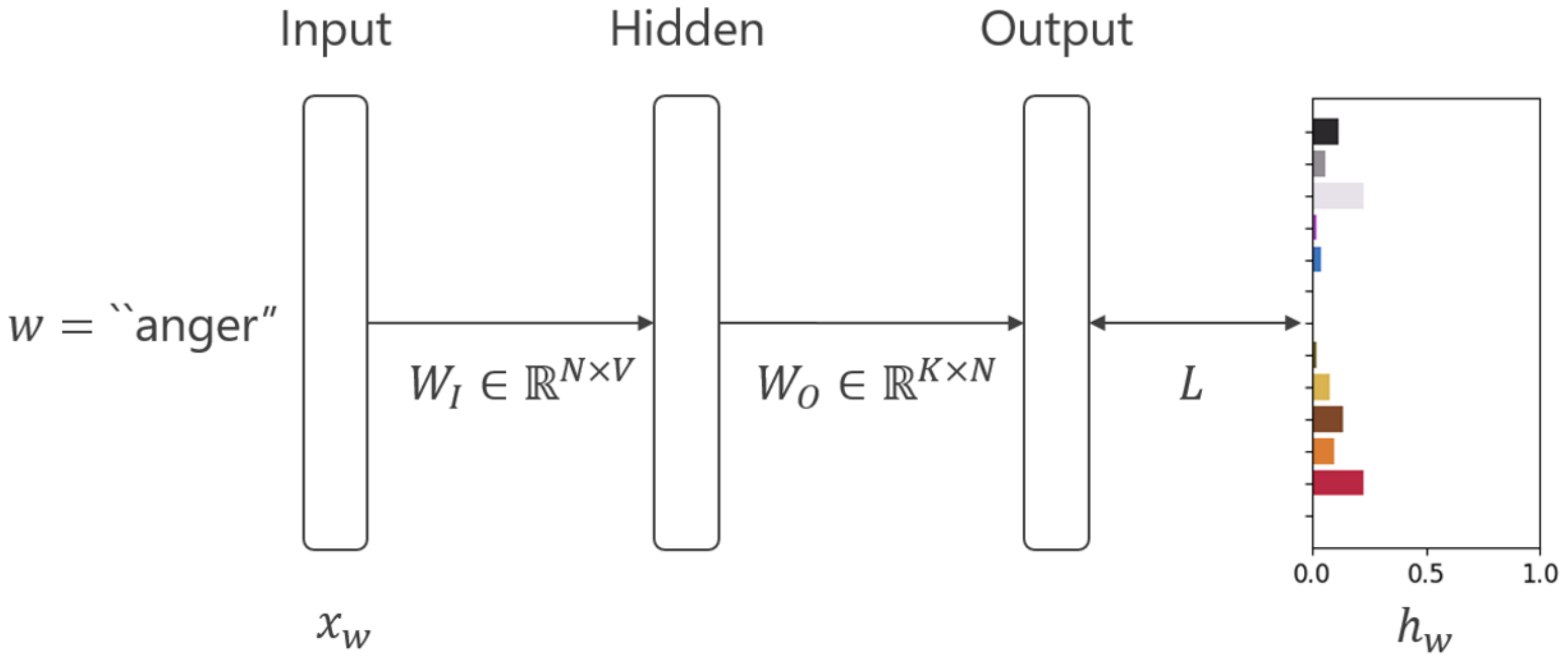}
	\caption{The neural network structure of the proposed word embedding method.\label{fig:cw2v}}
\end{center}
\end{figure}

Based on the analysis results of Section~\ref{sec:color-dist}, we attempt to create a word embedding grounded on word color usage using a neural network. 
Fig.~\ref{fig:cw2v} shows the network structure of the proposed word embedding method. 
This network is inspired by Sound-Word2vec~\cite{sw2v}. 
Specifically, this network is similar to an autoencoder or the skip-gram of word2vec. 
Its input $x_{w}$ is the $V$-dimensional one-hot representation of the input word $w$ and its output should be close to the $K$-dimensional vector $h_{w}$ representing the color histogram of $w$. 
The value $K=13$ is the number of bins of the color histogram. 
Accordingly, its output is not a $V$-dimensional one-hot vector but a $K$-dimensional vector representing a color histogram with $K$ bins. 
The $N$-dimensional output vector from the hidden layer is the semantic vector representation of input word $w$. 
\par

The neural network is trained to minimize the following loss function $L$:
\begin{equation}
  \label{eq:cw2v}
  L=\sum_w\|h_w-f(W_\mathrm{O}W_\mathrm{I}x_{w})\|,
\end{equation}
where the matrices $W_\mathrm{I}\in \mathbb{R}^{N\times V}$ and $W_\mathrm{O}\in\mathbb{R}^{K\times N}$ are the weight matrices to be trained. 
The function $f$ is softmax. 
The network tries to predict a color histogram $h_{w}$ through $W_\mathrm{O}$ and the softmax from $W_\mathrm{I}x_{w}$, which is the semantic vector representation of the word $w$. 
The matrix $W_\mathrm{I}$ is initialized as $\bar{W}_\mathrm{I}$ which is a word-embedding matrix trained by a word-embedding method, such as word2vec. 
The matrix $W_\mathrm{O}$ is initialized with random values.
\par
By updating $W_\mathrm{I}$ and $W_\mathrm{O}$ along with the minimization of $L$, semantic vectors become more different (similar) between words with different (similar) color usages. 
In other words, the proposed method retrains $W_\mathrm{I}$ from $\bar{W}_\mathrm{I}$ for giving more different (similar) semantic vectors for a pair of words, if they have different (similar) color usages. 
As noted above, typical word embedding methods, such as word2vec, often give similar semantic vectors even for a pair of antonyms. 
We, therefore, can expect that they have more different vectors if they have different color usages, as shown in Fig.~\ref{fig:concept}. 
More specifically, for a pair of antonyms, $x_{w_1}$ and $x_{w_2}$, we can expect $s_{w_1, w_2}-\bar{s}_{w_1, w_2}<0$, where $\bar{s}_{w_1, w_2} =\langle \bar{W}_\mathrm{I}x_{w_1}, \bar{W}_\mathrm{I}x_{w_2}\rangle / \|\bar{W}_\mathrm{I}x_{w_1}\|\|\bar{W}_\mathrm{I}x_{w_2}\|$ is the cosine similarity of the semantic vectors for $w_1$ and $w_2$ {\em without} color usage information, and $s_{w_1, w_2}=\langle W_\mathrm{I}x_{w_1}, W_\mathrm{I}x_{w_2}\rangle / \|W_\mathrm{I}x_{w_1}\|\|W_\mathrm{I}x_{w_2}\|$ is {\em with} color usage information. 
\par

\subsection{Evaluation\label{sec:evaluation}}
To analyze the positive or negative effect of using the color usage information of words, we focused pairs of synonyms and antonyms and evaluated their similarity before and after integrating the color usage information.
For the initial matrix $\bar{W}_\mathrm{I}$, GoogleNews-vectors (word2vec trained with Google News corpus) were used\footnote{The vocabulary assumed in GoogleNews-vectors is slightly different from that of our book title words. 
Specifically, 198 words in the book titles do not appear in GoogleNews-vectors. 
We, therefore, used 6,762-198 = 6,514 words in the experiment in this section. 
The data of GoogleNews-vectors is provided by https://code.\allowbreak google.com/archive/p/word2vec/.}.
We used synonym pairs and antonym pairs defined in Wordnet.

\subsubsection{Effect of Word Color on Similarity in Synonyms and Antonyms\label{sec:analy-cosine}}
\begin{figure}[t]
\begin{center}
 \includegraphics[width=\textwidth]{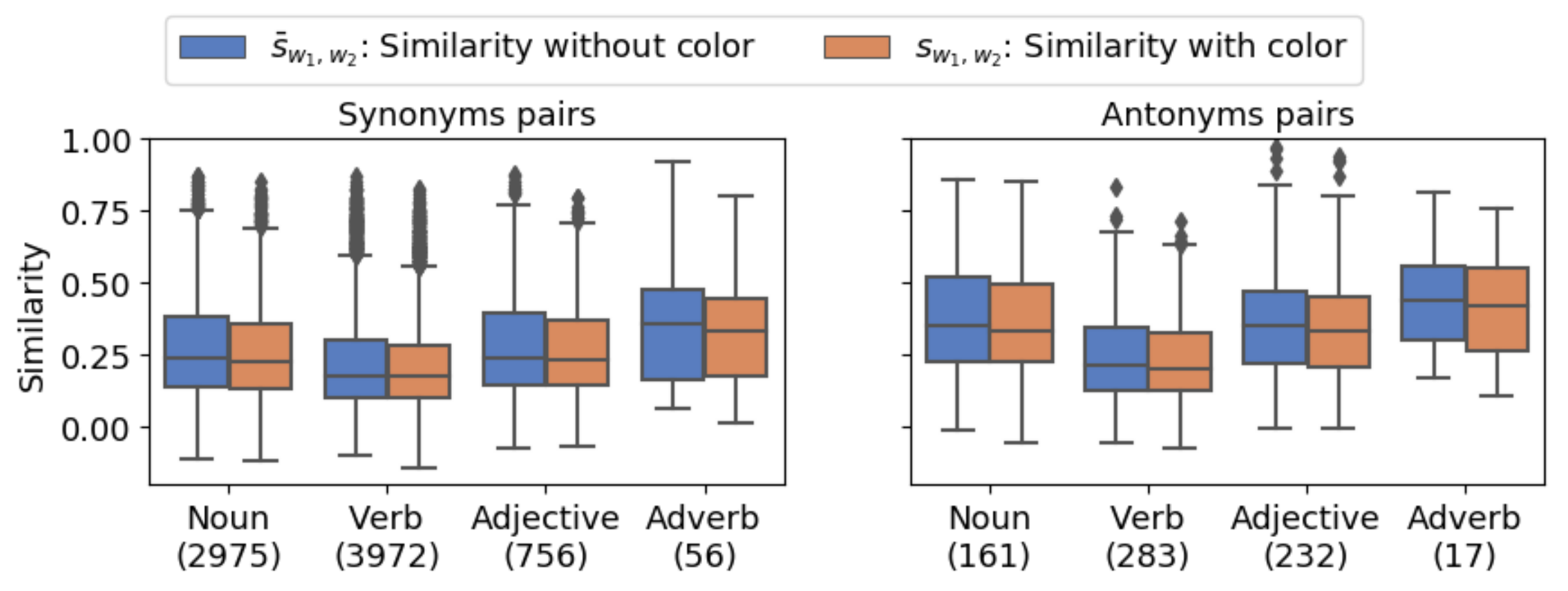}
\caption{Distributions of the similarity for synonym pairs (left) and antonym
 pairs (right). 
 \label{fig:box_cosine}} 
\end{center}
\end{figure}

As a quantitative evaluation, we observed the distributions of $\bar{s}_{w_1, w_2}$ and $s_{w_1, w_2}$ for all of the synonym pairs and antonym pairs. 
Fig.~\ref{fig:box_cosine} shows the similarity distributions with and without color. 
The highlight of this result is that the similarity between antonym pairs is {\em successfully} decreased when using color. 
This suggests that antonym pairs tend to have different color usages. 
However, this decrease in similarity also happens for the synonym pairs. 
Thus, we cannot assume that synonyms are always printed in similar colors.
Fig.~\ref{fig:box_cosine} also shows that the range of the similarity values for synonym pairs is similar to antonym pairs, even without color usage, that is, even by the original word2vec. 
(Ideally, synonym pairs should have higher similarities than antonym pairs.) 
This clearly shows the difficulty of the word embedding task. \par

\begin{figure}[t]
	\begin{center}
		\includegraphics[width=\textwidth]{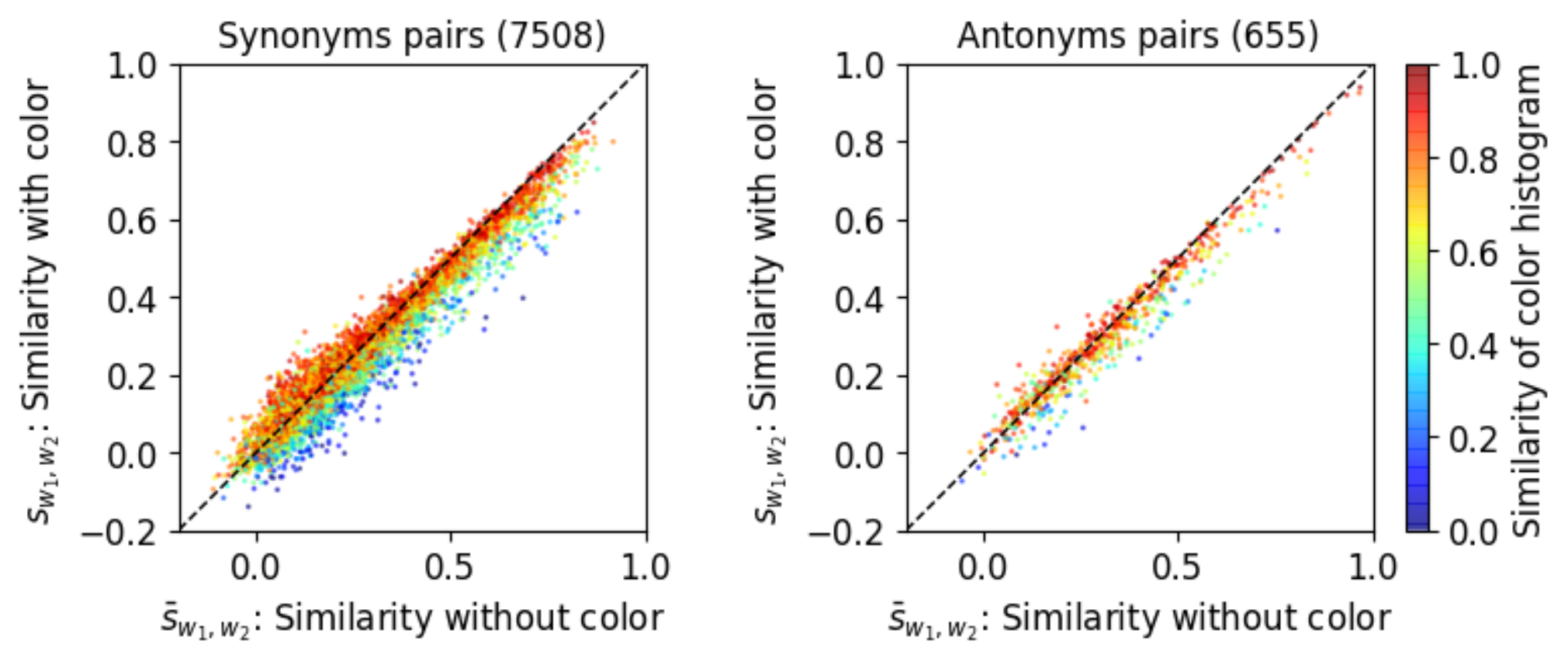}
		\caption{Similarity of the semantic vectors with or without
	 color information. Each dot corresponds to a pair of synonyms (left) or
	 antonyms (right).\label{fig:scat-synant}}
	\end{center}
\end{figure}

A more detailed analysis is shown in Fig.~\ref{fig:scat-synant}. 
This plot shows $(\bar{s}_{w_1, w_2},\allowbreak s_{w_1, w_2})$ for each synonym and antonym pair. 
Each point corresponds to a word pair. 
If a point is on the diagonal line, the corresponding word pair is not affected by the use of the color information in their semantic vector representation. 
The color of each point represents the similarity of the color information (i.e., the similarity of the color histograms of the paired words). 
It is observed that the proposed neural network works appropriately. 
In fact, pairs with different color usages (blue dots) are trained to have more different semantic vectors, that is, $\bar{s}_{w_1,w_2} > s_{w_1, w_2}$. 
In contrast, pairs with similar color usages (red dots) are trained to have more similar vectors, that is, $\bar{s}_{w_1,w_2} < s_{w_1, w_2}.$\par

Another observation of Fig.~\ref{fig:scat-synant} reveals that there are many blue points for synonym pairs and red points for antonym pairs.
Namely, synonym pairs often have different color usages and antonym pairs have similar usages. 
As we anticipated, the correlations between color usages and meanings for words are not always consistent and not always strongly correlated. 
This fact coincides with the quantitative evaluation result of Fig.~\ref{fig:box_cosine}.\par
%

\begin{figure}[t]
    \begin{center}
    \setlength\fboxsep{0pt}
    \setlength\fboxrule{0pt}
    \framebox[1\textwidth]{
    \includegraphics[width=0.85\textwidth]{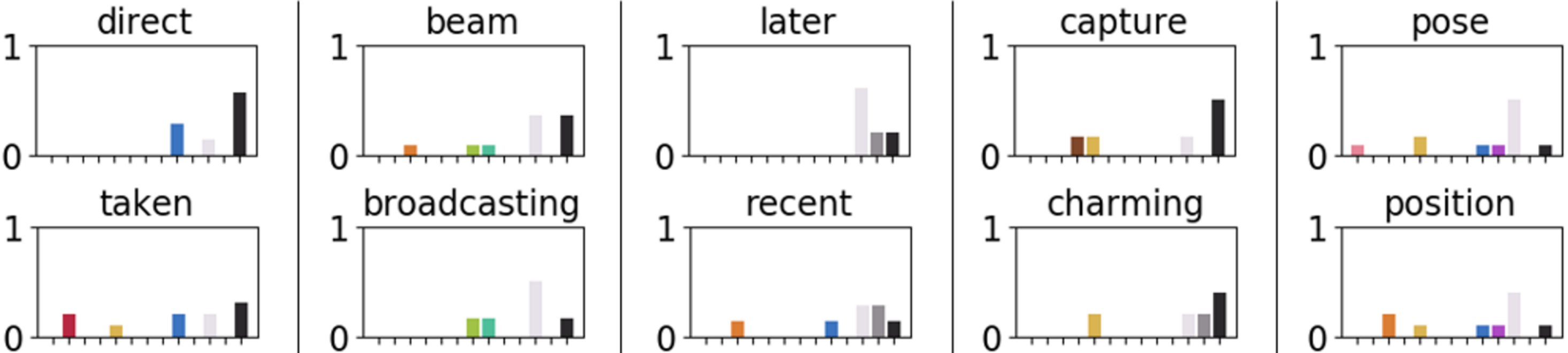}
    }
    \framebox[1\textwidth]{(a) Synonyms}
    \end{center}
    \begin{center}
    \setlength\fboxsep{0pt}
    \setlength\fboxrule{0pt}
    \framebox[1\textwidth]{
    \includegraphics[width=0.85\textwidth]{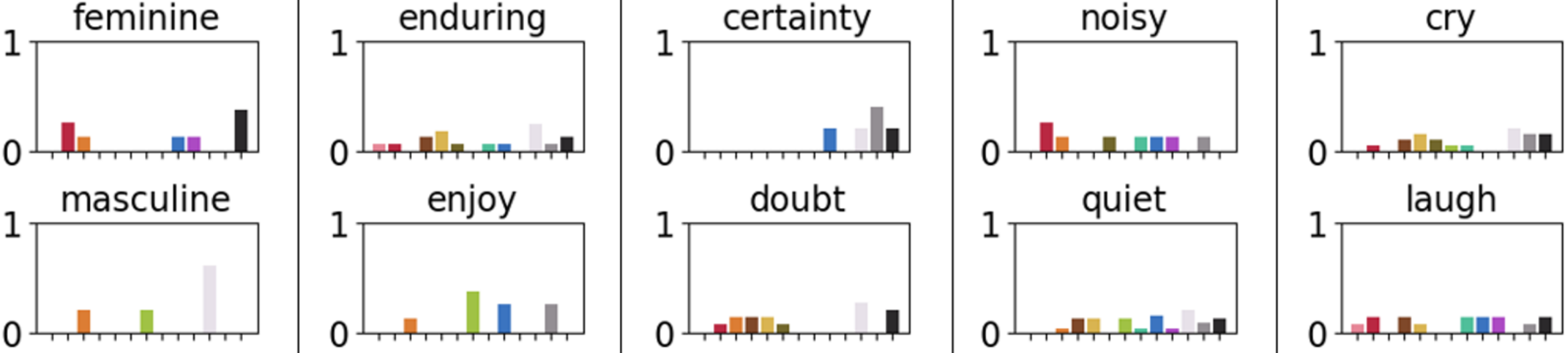}
    }
    \framebox[1\textwidth]{(b) Antonyms}
    \end{center}
    \caption{ 
    Example of (a) synonym pairs and (b) antonym pairs with improved similarity. Upper and lower histograms indicate pairs.
     }
    \label{fig:example-synant}
\end{figure}

Examples of synonyms and antonyms with an improved similarity between two words are shown in the Tables~\ref{tab:positive-syn} and~\ref{tab:positive-ant}. 
In Table~\ref{tab:positive-syn}, improvement is defined by a case where the similarity of the word embedding is increased for synonyms.
Conversely, in Table~\ref{tab:positive-ant}, improvement is defined by a case where the similarity is reduced for antonyms.
The lexnames in these tables indicate the category in which the two words are treated as synonyms or antonyms. 
From these, it can be seen that the word pairs used in verbs occupy the top of synonyms and word pairs of adjectives are common in antonyms. 
Next, Fig.~\ref{fig:example-synant} shows examples of color histograms of synonym and antonym word pairs. 
For the synonym pairs, the colors tend to be achromatic. 
This is because words, such as verbs, do not necessarily associate with color impressions and the visual characteristics of verbs tend to achromatic.
As a result, it can be seen that the word embedding with achromatic colors tended to be updated as more similar under the proposed method. 
In contrast, for antonyms, frequent chromatic differences between the word pairs can be confirmed. 
The correlation between the psychological impression of the color and the meaning of the word is strong. 
Thus, it can be said that for these words, namely adjectives, it is a good example of compensating for the differences in opposite words.

\begin{table}[t]
  \begin{minipage}[t]{.48\textwidth}
    \caption{Top 20 synonym pairs with similarity increase by color information.}
    \label{tab:positive-syn}
    \begin{center}
    \scalebox{0.74}[0.74]{
      \begin{tabular}{|c|c|c|c|}
\hline Word1 & Word2 & $s-\bar{s}$& Lexnames\\
\hline 
\hline direct & taken & 0.185 & \begin{tabular}{c}verb.motion\\verb.competition\end{tabular}\\ 
\hline broadcasting & spread & 0.176 & verb.communication\\
\hline hire & taken & 0.160 & verb.possession\\
\hline rolling & vagabond & 0.157 & verb.motion\\
\hline floating & vagabond & 0.156 & adj.all\\
\hline breed & covering & 0.149 & verb.contact\\
\hline fade & slice & 0.149 & noun.act\\
\hline gain & hitting & 0.146 & verb.motion\\
\hline lease & taken & 0.141 & verb.possession\\
\hline beam & broadcasting & 0.137 & verb.communication\\
\hline engage & taken & 0.133 & verb.possession\\
\hline shoot & taken & 0.130 & verb.communication\\
\hline conducting & direct & 0.130 & \begin{tabular}{c}verb.motion\\verb.creation\end{tabular}\\
\hline press & squeeze & 0.129 & verb.contact\\
\hline later & recent & 0.128 & adj.all\\
\hline affect & touching & 0.128 & verb.change\\
\hline demand & taken & 0.125 & verb.stative\\
\hline capture & charming & 0.124 & verb.emotion\\
\hline drop & spent & 0.124 & verb.possession\\
\hline pose & position & 0.122 & verb.contact\\
\hline 
\end{tabular}
}
    \end{center}
  \end{minipage}
  \hfill
  \begin{minipage}[t]{.48\textwidth}
    \caption{Top 20 antonym pairs with similarity decrease by color information.}
    \label{tab:positive-ant}
    \begin{center}
    \scalebox{0.74}[0.74]{
      \begin{tabular}{|c|c|c|c|}
\hline Word1 & Word2 & $s-\bar{s}$ & Lexnames\\
\hline 
\hline certain & uncertain & -0.192 & adj.all\\
\hline feminine & masculine & -0.184 & adj.all\\
\hline sit & standing & -0.158 & verb.contact\\
\hline nonviolent & violent & -0.137 & adj.all\\
\hline enduring & enjoy & -0.136 & verb.perception\\
\hline sit & stand & -0.135 & verb.contact\\
\hline correct & wrong & -0.133 & \begin{tabular}{c}adj.all\\verb.social\end{tabular}\\
\hline even & odd & -0.132 & adj.all\\
\hline spoken & written & -0.130 & adj.all\\
\hline certainty & doubt & -0.128 & noun.cognition\\
\hline laugh & weep & -0.127 & verb.body\\
\hline indoor & outdoor & -0.125 & adj.all\\
\hline lay & sit & -0.120 & verb.contact\\
\hline noisy & quiet & -0.119 & adj.all\\
\hline cry & laugh & -0.119 & verb.body\\
\hline host & parasite & -0.118 & noun.animal\\
\hline immortal & mortal & -0.118 & adj.all\\
\hline shrink & stretch & -0.118 & verb.change\\
\hline confident & shy & -0.117 & adj.all\\
\hline better & worse & -0.117 & adj.all\\
\hline 
\end{tabular}
}
\end{center}
\end{minipage}
\end{table}

\section{Conclusion}
To the authors' best knowledge, this paper is the first trial to analyze the relationship between the meaning and color of words, by using a data-driven (i.e., totally non-subjective) approach.
For this analysis, 358,712 word images were collected from the titles of 207,572 book cover images. 
First, we observed the words and the word categories with their color usages. 
Our analysis revealed the existence of words with {\em particular}, i.e. specific, color usage.  
For example, words relating to plants have more particular color usages than words for foods. 
On the other hand, we also found there are many words whose color usage is not particular. \par

We then observed the effect of the color usage information has on the word embedding, i.e. semantic vector representation of words. 
To do this, we developed a novel word-embedding method, which is inspired by Sound-Word2vec~\cite{sw2v}. 
The method modifies the word-embedding matrix given by word2vec to reflect color usages of individual words. 
Specifically, the similarity of the words with different color usages become more different and vise versa. 
We confirmed that the proposed method can provide a reasonable semantic vector representation.
By using antonym pairs and synonym pairs, we also confirmed that color information has ``positive'' or ``negative'' effects for the semantic vector representation. This is because there are synonym pairs with similar color usages but also synonyms with different color usages. 
The same situation happens to antonym pairs.\par

Since this is the first trial on a new research topic, we have many future works. 
The first and most important work is a more detailed analysis of the ``class'' of the words that have positive or negative effects from their color usage information. 
As we found through the analysis in this paper, there are two classes of antonyms; antonyms with different color usages and those with similar color usages. 
The former class will be improved by color information and the later will have degradation.
If we can find trends in these classes and use the proposed word-embedding method only for the former class, we can improve the performance of word embedding by color. 
This classification can also be done in a more general way, instead of focusing only on antonyms and synonyms. 
Of course, in addition to book covers, we can also introduce more word-color samples from the internet. 

Another future work is to consider a loss function specialized for color histogram in  the proposed word-embedding method. 
Currently, our loss function is the $L^2$ distance between the output and the 
actual color usage.  This means we treat 13 basic colors independently and thus the affinity 
between the basic colors is not considered. For example, red and orange are treated independently but a word often with red can be printed with orange. Moreover, the current results depend too much on the choice of 13 colors and their color space (e.g. CIELAB). Therefore, it will be meaningful to find a more suitable way for color usage representation.
%
%
\section*{Acknowledgment} This work was supported by JSPS KAKENHI Grant Number JP17H06100.
\bibliographystyle{splncs04}
\bibliography{ref} 
\end{document}